%% file: ijcai26.tex
\documentclass{article}
\usepackage{ijcai26}

\usepackage{times}
\usepackage{soul}
\usepackage{url}
\usepackage[hidelinks]{hyperref}
\usepackage[utf8]{inputenc}
\usepackage[small]{caption}
\usepackage{graphicx}
\usepackage{amsmath}
\usepackage{amsthm}
\usepackage{amssymb}
\usepackage{booktabs}
\usepackage{algorithm}
\usepackage{algorithmic}
\usepackage[switch]{lineno}
\usepackage[table,xcdraw]{xcolor}

\title{Uncertainty-aware Spatial-Frequency Registration and Fusion for Infrared and Visible Images}

\author{
    Xingyuan Li$^1$, 
    Haoyuan Xu$^1$, 
    Xingyue Zhu$^1$, 
    Jun Ma$^1$,
    Yang Zou$^2$, 
    Zhiying Jiang$^3$, 
    Jinyuan Liu$^1$\\
    \affiliations
    $^1$Dalian University of Technology;
    $^2$Northwestern Polytechnical University;
    $^3$Dalian Maritime University\\
    \emails
    xingyuan\_lxy@163.com, atlantis918@hotmail.com
}

\begin{document}
\twocolumn[{%
\renewcommand\twocolumn[1][]{#1}%
\maketitle
\vspace{-0.9cm}
\begin{center}
    \centering
    \captionsetup{type=figure}
    \vspace{-0.1in}
    \includegraphics[width=1\textwidth]{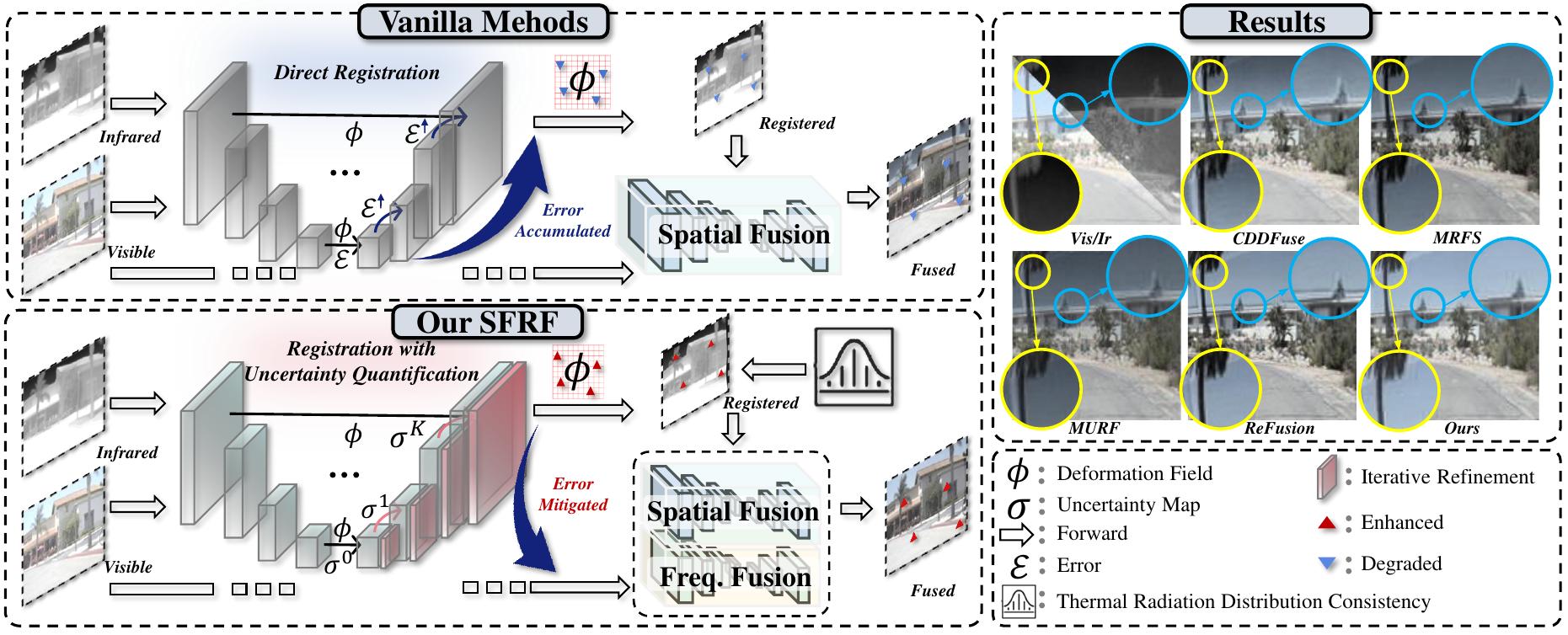}
    \vspace{-0.15in}
    \captionof{figure}{Vanilla methods often register the image directly, failing to account for the accumulated errors at coarse scales and severely degrading registration and fusion performance. They also neglect the specific frequency characteristics of infrared, which further diminishes the quality of the fusion. In contrast, our method addresses error accumulation and maintains thermal radiation distribution consistency through an uncertainty-aware spatial-frequency registration and fusion framework, ensuring more robust and accurate results.
    }
    
    \label{fig:teaser}
\end{center}%
}]


\input{sec/0_abstract}
\input{sec/1_intro}
\input{sec/2_related_works}
\input{sec/3_method}
\input{sec/4_experiment}

\newpage
\small
\bibliographystyle{named}
\bibliography{ijcai26}

\end{document}

%% file: sec/0_abstract.tex
\begin{abstract}
Infrared and Visible Image Fusion (IVIF) has shown promise in visual tasks under challenging environments, but fusion under unregistered conditions faces inherent misalignments. Current studies to solve them either predict the deformation parameters coarse-to-fine (i.e., coarse registration and fine registration) or estimate the deformation fields in multi-scales for registration. Though straightforward, they overlook the cumulative errors in registration, which contaminate the fusion stage and severely deteriorate the resulting images. We introduce the \textbf{S}patial-\textbf{F}requency \textbf{R}egistration and \textbf{F}usion (\textbf{SFRF}) framework, which incorporates uncertainty estimation and infrared thermal radiation distribution consistency into a unified pipeline to handle the error accumulation for robust registration and fusion across both spatial and frequency domains. Specifically, SFRF constructs a Multi-scale Iterative Registration (MIR) framework that iteratively refines the deformation field across scales, leveraging uncertainty estimation at each stage to mitigate error accumulation and enhance alignment accuracy dynamically. To ensure the accurate alignment of infrared thermal distributions during registration, thermal radiation distribution consistency is employed as a frequency-domain supervisory signal, promoting global consistency in the frequency domain. Based on the spatial-frequency alignment, SFRF further adopts a Dual-branch Spatial-Frequency Fusion (DSFF) module, which incorporates spatial geometric features and frequency distribution information to reconstruct visually appealing images. SFRF achieves impressive performance across diverse datasets.
\end{abstract}

%% file: sec/1_intro.tex
\section{Introduction}
\label{sec:intro}
Infrared and Visible Image Fusion (IVIF) is fundamental to integrating valuable information from different modalities in low-level vision~\cite{ma2019infrared,bai2025refusion}. Traditional IVIF methods
decompose images into multiple levels to extract information at different scales, typically relying on handcrafted feature extraction techniques and weighting rules. More recently, deep learning-based methods~\cite{guan2026domain,li2023text,xu2023murf,liu2024elegance,liu2024coconet} have demonstrated notable potential for downstream tasks such as object detection~\cite{liu2026bridging,wang2025efficient,liu2022target} and segmentation~\cite{kirillov2023segment,long2015fully,liu2023multi,11316813}, transforming the fusion process into a task-oriented approach. 

These approaches produce visually appealing results and hold unique industrial value for visual tasks in severe environments~\cite{xiao2026learning}. Yet, preserving the underlying image details and maintaining the integrity of modality-specific information is more challenging than it seems. The first challenge lies in handling infrared-specific spectral characteristics. As shown in Figure~\ref{fig:teaser}, most existing methods~\cite{liu2024promptfusion,li2024deep,wang2024improving,xiao2025incorporating,li2025difiisr,wang2022unsupervised,wang2020deepflash} are confined to spatial registration techniques, which fundamentally do not differ from visible image registration. This spatial-only focus neglects the unique spectral properties of infrared images, such as thermal radiation distributions, which are critical for achieving effective infrared and visible image fusion~\cite{zhao2023cddfuse,bai2025task}. As a result, this oversight often degrades the fusion quality, especially in scenarios where preserving modality-specific information is essential.

The second challenge lies in the inherent uncertainty in estimating the deformation field during registration, whether in coarse-to-fine (i.e., coarse and fine registration)~\cite{xu2023murf} or single-stage multi-scale registration frameworks~\cite{wang2024improving,wang2022unsupervised,huang2022reconet}. Due to the complex nature of infrared-visible image registration, including multi-scale spatial deformations and cross-modal disparities, predicting a precise deformation field is inherently uncertain~\cite{beauchemin1995computation,decarlo2000optical,brox2010large}. This uncertainty is further exacerbated by the single-pass prediction strategies commonly used in registration frameworks, which propagate and amplify errors. 


As a response, we introduce the \textbf{S}patial-\textbf{F}requency \textbf{R}egistration and \textbf{F}usion (SFRF) framework to address error accumulation while preserving thermal radiation distribution consistency. SFRF quantifies uncertainty at each scale, mitigating penalties in uncertain areas and discouraging excessively large uncertainties through a Multi-scale Iterative Registration (MIR) framework, which helps reduce accumulated errors during registration. To preserve infrared-specific characteristics, SFRF incorporates thermal radiation distribution consistency, guiding infrared image registration for improved fusion results. Additionally, SFRF employs a Dual-branch Spatial-Frequency Fusion (DSFF) module to capture spatial geometric features and frequency distribution information, thereby reconstructing visually appealing images. Our contributions are as follows:
\begin{itemize}
\item  We propose \textbf{S}patial-\textbf{F}requency \textbf{R}egistration and \textbf{F}usion (\textbf{SFRF}) framework to address the misalignments in fusion under unregistered conditions. 

\item In the registration phase, we introduce the Multi-scale Iterative Registration (MIR) framework to mitigate accumulated spatial-domain errors and adopt thermal-radiation distribution consistency as a frequency-domain supervisory signal, ensuring precise alignment of infrared thermal distributions and promoting global frequency-domain consistency.
\item In the fusion phase, we present the Dual-branch Spatial-Frequency Fusion (DSFF) module to reconstruct visually appealing and perceptually coherent images by integrating spatial geometric features and frequency distribution information.
\end{itemize}

%% file: sec/2_related_works.tex
\begin{figure*}[!pth]
    \centering
    \includegraphics[width=1\linewidth]{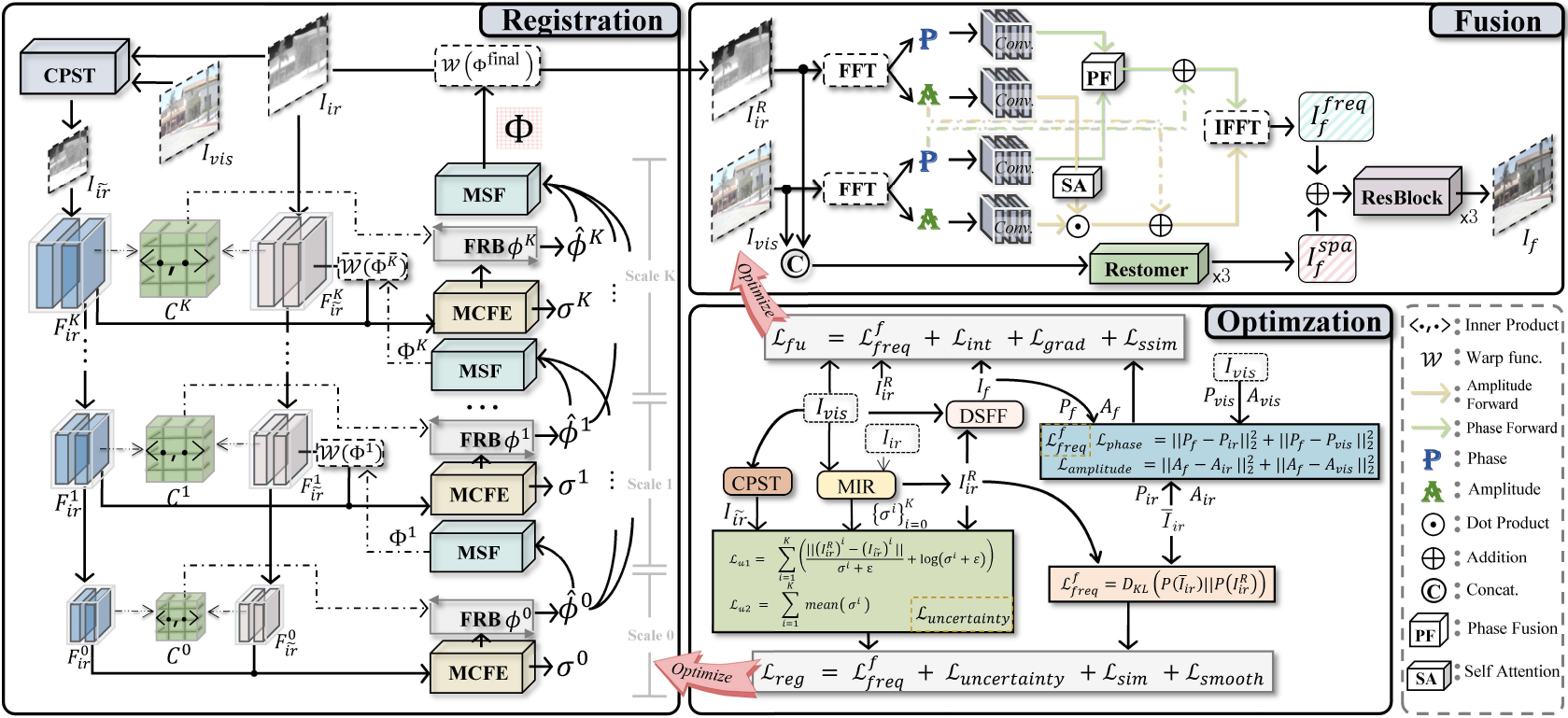}
    \caption{Overview of SFRF. }
    \label{fig:pipeline}
\end{figure*}
\section{Related Work}
\label{sec:related}

\subsection{Multi-Modal Image Registration}
Traditional multi-modal image registration methods often fall into intensity-based, feature-based, or transformation-based categories. While intensity-based approaches compare pixel intensities across modalities 
\cite{liu2025toward}
, feature-based methods rely on extracting distinctive features, and transformation-based methods estimate geometric transformations. Deep learning-based methods focus on optimizing the parameters for predicted affine transformations or deformation fields. For instance, Arar et al.\cite{10} introduced a geometrically preserving image transformation network. Ye et al.\cite{ye2022multiscale} stacked deep networks to map image pairs to transformation parameters for improved accuracy. Nevertheless, multi-modal registration typically involves multi-scale deformation field estimation. Varying texture or lighting across scales introduces uncertainty in affine parameters \cite{decarlo2000optical,10849824}, which can amplify errors through iterative optimization. Moreover, most methods focus on spatial alignment alone, neglecting the spectral distribution in infrared images, which is crucial for high-quality fusion results.

\subsection{Multi-Modal Image Fusion}
Autoencoder-based fusion methods learn low-dimensional representations to fuse different modalities \cite{ren2021infrared}, while CNN-based approaches typically involve either feature extraction with weight generation \cite{liu2025dcevo} or end-to-end learning \cite{hou2020vif}. GAN-based methods implicitly integrate feature extraction, feature fusion, and image reconstruction \cite{li2020attentionfgan,li2026unifusion,xiao2025occlusion}. Although frequency-domain information has proven beneficial for fusion, many existing methods overlook critical amplitude–phase differences, leading to the loss of modality-specific features. For instance, Hu et al.~\cite{hu2024sfdfusion} treated the amplitude and phase of infrared and visible images in a unified manner, simply concatenating them for convolutional fusion. Huang et al.\cite{huang2022reconet} proposed a recurrent correction network with a deformation module and attention mechanism for slight misalignments. 

%% file: sec/3_method.tex
\section{Method}
\label{sec:method}
As shown in Figure~\ref{fig:pipeline}, given a pair of unaligned infrared and visible images \(I_{ir}\) and \(I_{vis}\), respectively, our proposed \textbf{S}patial-\textbf{F}requency \textbf{R}egistration and \textbf{F}usion (\textbf{SFRF}) framework incorporates uncertainty estimation and infrared thermal radiation distribution consistency into a unified pipeline to handle the error accumulation for robust registration and fusion across both spatial and frequency domains. 


\subsection{Multi-scale Iterative Registration (MIR)}
As shown in Figure~\ref{fig:pipeline} (Registration), given the input distorted infrared image \(I_{ir}\) and the pseudo-infrared image \(I_{\widetilde{ir}}\) generated from the visible image \(I_{vis}\) through a frozen CPST~\cite{wang2022unsupervised} module, an encoder comprising LK-CNN blocks~\cite{ding2024unireplknet} is employed to extract multi-scale feature representations of both \(I_{ir}\) and \(I_{\widetilde{ir}}\). Specifically, for each image \(I\), the encoder outputs \(K\) levels of feature maps from the coarsest scale \(i=0\) to the finest scale \(i=K\): \(\mathbf{F}_{ir} = \{F_{ir}^0, F_{ir}^1, ..., F_{ir}^K\}\), and \(\mathbf{F}_{\widetilde{ir}} = \{F_{\widetilde{ir}}^{0}F_{\widetilde{ir}}^{1}, ..., F_{\widetilde{ir}}^{K}\}\), where \(F_{ir}^i\) and \(F_{\widetilde{ir}}^{i}\) represent the feature maps at scale \(i\). To capture the feature-level similarity between \(I_{ir}\) and \(I_{\widetilde{ir}}\), we construct a correlation map \(C^i \;\in\;\mathbb{R}^{B\times 1\times H\times W}\) at each scale as \(C^i = \frac{F_{ir}^i \cdot (F_{\widetilde{ir}}^{i})^T}{\|F_{ir}^i\| \|F_{\widetilde{ir}}^{i}\|}\), for \( i \in \{0, 1, ..., K\}\), where each spatial location \((y,x)\) in \(C^i\) encodes the mean similarity of \(\bigl(F_{ir}^i\bigr)_{(y,x)}\) against all positions in \(F_{\widetilde{ir}}^{i}\). 

\noindent\textbf{Monte Carlo sampling-based field estimation (MCFE).}
At the coarsest scale (\(i = 0\)), an initial coarse deformation field is predicted via a Monte Carlo sampling-based field estimation (MCFE) module that incorporates dropout to enable uncertainty quantification. Let \(\mathcal{D}\) denote the set of dropout-enabled forward passes through the MCFE module, and for each forward pass \(n \in \{1, \dots, N\}\), the predicted deformation field \(\phi^0_n\) is given by
\(\phi^0_n = \mathcal{F}\left(F_{ir}^0, F_{\widetilde{ir}}^{0}; \mathcal{D}_n\right)\), where \(\mathcal{F}\) is the MCFE module applied to the feature maps \(F_{ir}^0\) and \(F_{\widetilde{ir}}^{0}\) at scale 0, with \(\mathcal{D}_n\) denoting the \(n\)-th stochastic dropout mask applied during the forward pass. The coarse deformation field \(\phi^0\) is then computed as the mean of the \(N\) predictions (\(N\) = 10 in our experiments), and the uncertainty map \(\sigma^0\) is estimated based on the variance, as \(\phi^0 = \frac{1}{N}\sum_{n=1}^{N} \phi^0_n\) and \(\sigma^0 = \frac{1}{N} \sum_{n=1}^{N} \left(\phi^0_n - \phi^0\right)^2\).

\noindent\textbf{GRU-based field refinement block (FRB).}
Due to the complex nature of infrared-visible image registration, predicting a precise deformation field is inherently uncertain
~\cite{brox2010large}
. This uncertainty is further exacerbated by the single-pass prediction, leading to the propagation and amplification of errors. MIR iteratively refines the initial deformation field at each scale via a field refinement block (FRB) using a GRU-based update mechanism guided by the corresponding correlation map \(C\) to mitigate this error amplification. Define the FRB update operator \(\mathcal{G}\) by \(\mathcal{G}\left(\phi, h\right) = \operatorname{Conv}\!\Bigl( \operatorname{GRU}\Bigl(\operatorname{concat}\bigl(\phi, F, C\bigr),\, h\Bigr) \Bigr)\), where \(\operatorname{concat}(\cdot)\) represents channel-wise concatenation, \(\operatorname{GRU}(\cdot,\cdot)\) is the GRU cell that maps the concatenated input and previous hidden state \(h\) to an updated hidden state, and \(\operatorname{Conv}(\cdot)\) is a 3\(\times\)3 convolution operation that projects the hidden state into the deformation field space. The iterative refinement process over \(T\) steps is then expressed as:
\begin{equation}
\phi_{(t+1)} \;=\; \phi_{(t)} \;+\; \mathcal{G}\Bigl(\phi_{(t)},\,h_{(t)}\Bigr), \quad t=0,\dots,T-1,
\end{equation}
\noindent where the \(h_{(t)}\) denotes the hidden state at iteration \(t\) (initialized with zeros), and the summation represents a residual accumulation of the GRU updates (\(T\) = 3 in our experiments). The initial deformation field \(\phi^0\) is then iteratively refined through the FRB to progressively enhance the alignment, resulting in the refined deformation field at the coarsest scale as \(\hat{\phi}^0 = \phi^0_{(T)}\), where \(\phi^0_{(T)}\) denotes the deformation field after \(T\) refinement iterations.

\noindent\textbf{Multi-scale field fusion (MSF).}
MIR utilizes the Multi-scale Field Fusion (MSF) to mitigate the error accumulation, which refines hierarchical deformation fields by ensuring spatial consistency and uncertainty-adaptive weighting before propagating transformations to finer scales. 

Given a refined deformation field \(\hat{\phi}^k\) at scale \(k\), \(k<K\), we first upsample all previously estimated deformation fields \(\{\hat{\phi}^i\}_{i=0}^{k-1}\) to the resolution of scale \(k\) by \(\widetilde{\phi}^{(i\to k)}
=\mathrm{Up}\!\bigl(\hat{\phi}^i,\,s_k\bigr), \forall i < k\), where \(s_k\) is the appropriate upsampling factor, and \(\mathrm{Up}(\cdot)\) represents bilinear interpolation. However, due to potential spatial misalignment across scales, directly combining these deformation fields can lead to accumulated registration errors. To alleviate this, MSF embodies a correlation-based alignment step to compute a correlation map that measures the similarity between each upsampled deformation field and \(\hat{\phi}^k\), formally \(\mathcal{R}^{(i,k)}=\mathrm{Corr}\!\Bigl(\widetilde{\phi}^{(i\to k)},\, \hat{\phi}^k\Bigr)\), where \(\mathrm{Corr}(\cdot,\cdot)\) is a pixel-wise normalized correlation function that estimates feature alignment quality. A small network (i.e., a \(3\times3\) convolution + sigmoid activation) then modulates \(\widetilde{\phi}^{(i\to k)}\) as:
\begin{equation}
\widehat{\phi}^{(i\to k)}=\widetilde{\phi}^{(i\to k)} \odot \sigma\bigl(\mathcal{R}^{(i,k)}\bigr),
\end{equation}
\noindent where \(\sigma(\cdot)\) is the sigmoid function, and \(\odot\) denotes elementwise multiplication. This spatial adaptation step ensures that \(\widehat{\phi}^{(i\to k)}\) better matches the finer-scale field \(\hat{\phi}^k\), preventing naive field summation from amplifying spatial misalignments.

Beyond spatial alignment, MSF leverages uncertainty estimation to assign reliability-aware weights to each scale’s deformation field. Given an uncertainty map \(\sigma^i\) associated with each \(\hat{\phi}^i\), we define a confidence-based weighting scheme as \( \omega^{(i)}=\frac{\exp\!\bigl(-\,\beta\,\sigma^i \odot \mathcal{R}^{(i,k)}\bigr)}{\sum_{j=0}^{k}\,\exp\!\bigl(-\,\beta\,\sigma^j \odot \mathcal{R}^{(j,k)}\bigr)}\), where \(\beta>0\) is a hyperparameter controlling the confidence scaling. This weighting scheme suppresses unreliable deformation fields while retaining informative ones. The aligned and confidence-weighted multi-scale flows are then fused as:
\begin{equation}
\widetilde{\Phi}^k=\sum_{i=0}^{k-1} \omega^{(i)}\,\widehat{\phi}^{(i\to k)}+\omega^{(k)}\,\hat{\phi}^k,
\end{equation}
\noindent where \(\widetilde{\Phi}^k\) denotes the fused deformation field at scale \(k\). Finally, for additional regularization, we apply a smoothing operation using velocity field integration as \(\Phi^k= \mathrm{VecInt}\bigl(\widetilde{\Phi}^k,\,n\bigr)\), where \(\mathrm{VecInt}(\cdot, n)\) denotes an \(n\)-step scaling-and-squaring integration.

The fused deformation field \(\Phi^k\) serves as a guidance transformation, applied to warp the infrared feature map at the next scale \(F_{ir}^{k+1}\) as \(F_{ir}^{k+1}=\mathcal{T}\!\bigl(F_{ir}^{k+1},\;\Phi^k\bigr)\), where \(\mathcal{T}(\cdot,\cdot)\) denotes spatial transformation using the fused field. This pre-alignment step mitigates large-scale errors, enabling finer-scale registration to focus on local refinements. Then, an initial flow \(\phi^{\,k+1}\) and an uncertainty map \(\sigma^{k+1}\) are again predicted via the MCFE module, followed by iterative refinement through the FRB. This yields a refined field \(\hat{\phi}^{\,k+1}\). The same MSF procedure then fuses \(\hat{\phi}^{\,k+1}\) with all coarser fields \(\{\hat{\phi}^0,\dots,\hat{\phi}^k\}\), producing \(\Phi^{k+1}\). By iterating from coarse (\(k=0\)) to fine (\(k=K\)) scales in this manner, MIR effectively reduces error accumulation and achieves a robust multi-scale registration. The final deformation field \(\Phi^{final}\) fuses all fields \(\{\hat{\phi}^0,\dots,\hat{\phi}^K\}\) and warps the distorted infrared image \(I_{ir}\) into the registered infrared image \(I_{ir}^R\).

\noindent \textbf{Loss function for registration.} 
Given the registered image \(I_{ir}^R\) and its undistorted ground truth \(\overline{I}_{ir}\), we compute their 2D Fourier transforms \(F_R = \mathcal{F}(I_{ir}^R)\) and \(F_{GT} = \mathcal{F}(\overline{I}_{ir})\), and extract the magnitude spectra \(M_R = |F_R|\) and \(M_{GT} = |F_{GT}|\). These are normalized into probability distributions by \(P(I) = \frac{A(I)}{\sum_{u,v} A(I)(u,v)}\), where \(A(I)\) denotes the amplitude of \(I\) in the frequency domain. The frequency consistency loss \(\mathcal{L}^{r}_{freq}\) is computed by measuring the frequency-domain similarity between the normalized distributions using the Kullback-Leibler (KL) divergence, formally:
\begin{equation}
\begin{aligned}
\mathcal{L}^{r}_{freq} 
&= D_{KL}\!\Bigl( P\bigl(\overline{I}_{ir}\bigr) \,\Big\|\, P\bigl(I_{ir}^R\bigr) \Bigr) \\[1mm]
&= \sum_{u,v} P\bigl(\overline{I}_{ir}\bigr)(u,v)\,\log\!\left(\frac{P\bigl(\overline{I}_{ir}\bigr)(u,v)}{P\bigl(I_{ir}^R\bigr)(u,v)+\varepsilon}\right),
\end{aligned}
\end{equation}
\noindent where \(\varepsilon>0\) is a small constant to avoid division by zero. 

In the spatial domain, the proposed MIR leverages an uncertainty loss to handle ambiguous regions robustly. Recall that the Monte Carlo sampling-based field estimation (MCFE) module produces a field \(\phi^i\) and an uncertainty map \(\sigma^i\) at each scale \(i\), for \( i \in \{0, 1, ..., K\}\). MIR utilizes these multi-scale uncertainty maps to incorporate two complementary terms that mitigate penalties in uncertain areas while discouraging excessively large uncertainties. First, an uncertainty-weighted reconstruction term:
\begin{equation}
\mathcal{L}_{u1}
=\;\sum_{i=1}^{K}
\Bigl(
\frac{\bigl\|\,(I_{ir}^R)^i - (I_{\widetilde{ir}})^i\bigr\|}{\sigma^i+\varepsilon}
\;+\;
\log(\sigma^i+\varepsilon)
\Bigr),
\end{equation}
\noindent down weights residuals in low-confidence regions while penalizing excessively large \(\sigma^i\). Second, an uncertainty regularization term \(\mathcal{L}_{u2}
=\;\sum_{i=1}^{K}\mathrm{mean}\bigl(\sigma^i\bigr)\), prevents inflating \(\sigma^i\) globally. The combined uncertainty loss \(\mathcal{L}_{uncertainty}\) is computed by the sum of \(\mathcal{L}_{u1}\) and \(\mathcal{L}_{u2}\). By adaptively reducing penalties in uncertain areas and imposing global constraints on uncertainty, MIR achieves robust alignment in ambiguous regions without resorting to trivially large uncertainty everywhere. The uncertainty loss \(\mathcal{L}_{uncertainty}\) together with the similarity loss \(\mathcal{L}_{sim}\) and the smooth loss \(\mathcal{L}_{smooth}\) forms the spatial consistency loss \(\mathcal{L}^{r}_{spa}\). The total registration loss \(\mathcal{L}_{reg}\) is as \(\mathcal{L}_{reg} = \mathcal{L}^{r}_{freq} + \mathcal{L}^{r}_{spa}\).

\subsection{Dual-branch Spatial-Frequency Fusion}
After registration, SFRF employs a Dual-branch Spatial-Frequency Fusion (DSFF) to fuse the infrared and visible images by leveraging their complementary spatial and frequency information, as shown in Figure~\ref{fig:pipeline} (Fusion).

\noindent\textbf{Frequency domain fusion.} 
DSFF first apply the Fast Fourier Transform (FFT) to the registered infrared image \(I_{ir}^R\) and the visible image \(I_{vis}\), obtaining their respective amplitude (\(P\)) and phase (\(A\)) components as \(P_{ir}, A_{ir} = \operatorname{FFT}(I_{ir}^R)\), \(P_{vis}, A_{vis} = \operatorname{FFT}(I_{vis})\). For each component, two 1 \(\times\) 1 convolutional layers are applied to extract features, denoted as \(F_{ir}^P\), \(F_{vis}^P\), \(F_{ir}^A\), \(F_{vis}^A\). 

To fuse the \(\emph{phase}\) components, the DSFF block adopts a phase fusion module (\(\mathrm{PF}\)) equipped with a heat-aware adaptive weighting mechanism. Let \(\mathrm{PF}\) generate an attention map \(a \in [0,1]\) that adaptively selects between infrared and visible phase features, ensuring thermal information dominates in high-temperature regions while visible phase cues prevail in background areas. The final phase fusion \(P_f\) is then computed via:
\begin{equation}
P_f = a*F_{ir}^P + (1-a)*F_{vis}^P + P_{vis}.
\end{equation}
Similarly, \(\emph{amplitude}\) fusion is achieved by a self-attention (SA) mechanism applied to the extracted amplitude features \(F_{ir}^A\) and \(F_{vis}^A\). Let \(\mathrm{SA}(\cdot)\) compute attention weights that highlight critical thermal structures in \(F_{ir}^A\). We then obtain the fused amplitude \(A_f\) via
\begin{equation}
A_f 
\;=\; 
\mathrm{SA}\bigl(F_{ir}^A\bigr)
\,\odot\,F_{vis}^A 
\;+\;
A_{ir}.
\end{equation}

This amplitude fusion mechanism highlights critical thermal structures, emphasizing regions with strong infrared responses. Finally, the fused frequency-domain image \(I_f^{freq}\) is reconstructed through the inverse Fourier transform as \(I_{f}^{freq} = \text{IFFT}(P_f, A_f)\), which integrates thermal saliency from \(I_{ir}^R\) while retaining the structural integrity of \(I_{vis}\) in the frequency domain.

\noindent\textbf{Spatial domain fusion.} 
In addition to frequency-domain fusion, DSFF also performs spatial fusion on the registered infrared image \(I_{ir}^R\) and the visible image \(I_{vis}\). Specifically, we concatenate them along the channel dimension as \(\mathbf{X}= \mathrm{concat}\bigl(I_{ir}^R,\;I_{vis}\bigr)\). The spatial fusion result is computed by \(I_{f}^{spa}= \Phi_{spa}\!\bigl(\mathbf{X}\bigr)\), where \(\Phi_{spa}\) denote the spatial fusion operator composed of three Restormer blocks~\cite{zamir2022restormer}. Each Restormer block refines the concatenated features by exploiting long-range dependencies via self-attention, thereby balancing detail preservation (from \(I_{vis}\)) and thermal saliency (from \(I_{ir}^R\)) in the spatial domain.

Finally, the spatial fusion result \(I_{f}^{spa}\) and the frequency fusion result \(I_{f}^{freq}\) are combined and further refined by a stack of ResBlocks~\cite{he2016deep} to produce the final fused output as \(I_{f}^{final}= \Phi_{res}\!\Bigl(\mathrm{concat}\bigl(I_{f}^{spa},\, I_{f}^{freq}\bigr)\Bigr)\), where \(\Phi_{res}\) denotes a sequence of four residual blocks, ensuring that the fused image retains spatially accurate details and frequency-consistent thermal cues. 

\noindent \textbf{Loss function for fusion.}
To encourage the final fused image \(I_{f}^{final}\) to preserve both the infrared-specific thermal cues and the visible structural fidelity in the frequency domain, we impose two separate constraints on its \emph{phase} and \emph{amplitude}. Let \(P_{f}, A_{{f}}\) be the phase and amplitude of the final fusion result \(I_{f}^{final}\), while \(\{P_{ir}, A_{ir}\}\) and \(\{P_{vis}, A_{vis}\}\) are those of the ground-truth infrared and visible images. We define the frequency loss \(\mathcal{L}^{f}_{freq}\) as \(\mathcal{L}^{f}_{freq}=
\mathcal{L}_{phase}+\mathcal{L}_{amplitude}\), where \(\mathcal{L}_{phase}=\bigl\|P_{{f}} - P_{ir}\bigr\|_{2}^{2}+\bigl\|
P_{{f}} - P_{vis}
\bigr\|_{2}^{2}\) and \(\mathcal{L}_{amplitude}=\bigl\|
A_{{f}} - A_{ir}\bigr\|_{2}^{2}+\bigl\|
A_{{f}} - A_{vis}\bigr\|_{2}^{2}\). This frequency loss ensures that the fuse result balances between infrared and visible to maintain thermal characteristics while preserving overall structural integrity. The total fusion loss \(\mathcal{L}_{fu}\) consists of the frequency loss \(\mathcal{L}^{f}_{freq}\) and the spatial loss \(\mathcal{L}^{f}_{spa}\), formally \(\mathcal{L}_{fu} = \mathcal{L}^{f}_{freq} + \mathcal{L}^{f}_{spa},\) where the \(\mathcal{L}^{f}_{spa}\) comprising the intensity loss \(\mathcal{L}_{int}\), gradient loss \(\mathcal{L}_{grad}\) and the SSIM loss \(\mathcal{L}_{ssim}\).


%% file: sec/4_experiment.tex
\begin{table*}[!t]
\centering
  \scriptsize
  \renewcommand\arraystretch{1.1} 
  \setlength{\tabcolsep}{0.8mm}
  \caption{\textbf{Quantitative comparison of image registration.} The best results are in \textbf{bold}, while the second-best results are \underline{underlined}.}
  \resizebox{\textwidth}{!}{
  \begin{tabular}{lc|ccc|ccc|ccc|ccc}
    \toprule
    \multicolumn{2}{c|}{\cellcolor{gray!13}}     & \multicolumn{3}{c|}{\cellcolor{gray!14}RoadScene} & \multicolumn{3}{c|}{\cellcolor{gray!14}MSRS} & \multicolumn{3}{c|}{\cellcolor{gray!14}\(\text{M}^{3}\text{FD}\)} & \multicolumn{3}{c}{\cellcolor{gray!14}TNO}  \\
    \multicolumn{2}{c|}{\cellcolor{gray!14}Methods} & \cellcolor{gray!14}NCC$\uparrow$ & \cellcolor{gray!14}RMSE $\downarrow$ & \cellcolor{gray!14}MEE $\downarrow$ & \cellcolor{gray!14}NCC $\uparrow$ & \cellcolor{gray!14}RMSE $\downarrow$ & \cellcolor{gray!14}MEE $\downarrow$ & \cellcolor{gray!14}NCC $\uparrow$ & \cellcolor{gray!14}RMSE $\downarrow$ & \cellcolor{gray!14}MEE $\downarrow$ & \cellcolor{gray!14}NCC $\uparrow$ & \cellcolor{gray!14}RMSE $\downarrow$ & \cellcolor{gray!14}MEE $\downarrow$  \\
    \midrule
    GLU-Net~\cite{truong2020glu}  & CVPR'20   & 0.805  & \underline{6.629}  & \underline{16.133}  & 0.551  & 6.318 & 10.438  & 0.729  & 6.110 & 9.983 & 0.471 & \textbf{7.757} & \textbf{29.000}    \\
    CGRP~\cite{wang2022unsupervised}  & IJCAI'22    & 0.907  & 7.883  & 36.467  & 0.394  & 6.703  & 17.539  & 0.885  & 6.186 & 13.970 & 0.746 & 8.400 & 45.250   \\
    SuperFusion~\cite{tang2022superfusion} & JAS'22 & 0.923  & \textbf{6.011}  & \textbf{13.267}  & \underline{0.571}  & \underline{5.916}  & \textbf{7.978}  & \underline{0.926}  & \underline{5.997} & \underline{5.290} & \underline{0.869} & 8.122 & 39.250    \\
    MURF~\cite{xu2023murf}  & TPAMI'23     & 0.919  & 7.437  & 27.767  & 0.562  & 6.427  & 12.533  & 0.897  & 6.043  & 8.580  & 0.779 & 8.115 & 41.000    \\
    IMF~\cite{wang2024improving}   & TCSVT'24    & \underline{0.926}  & 7.613  & 33.333  & 0.409  & 6.756  & 18.966  & 0.901  & 5.998 & 12.087 & 0.749 & 8.138 & 44.750   \\
    BSAFusion~\cite{li2024bsafusion}   & AAAI'25    & 0.809  & 7.782  & 31.933  & 0.423  & 6.335  & 11.989  & 0.777  & 6.476 & 16.328 & 0.696 & 8.257 & 42.500    \\
    \midrule
    \textbf{Ours}   &\textbf{-} & \textbf{0.947} & 7.370 & 25.400 & \textbf{0.573} & \textbf{5.818} & \underline{9.270} & \textbf{0.948} & \textbf{5.073} & \textbf{5.043} & \textbf{0.873} & \underline{8.053} & \underline{38.500}  \\
    \bottomrule
  \end{tabular}
   }
  \label{tab:reg_quantitative_results}%
\end{table*}

\begin{figure*}[!t]
 \centering
 \includegraphics[width=1\textwidth]{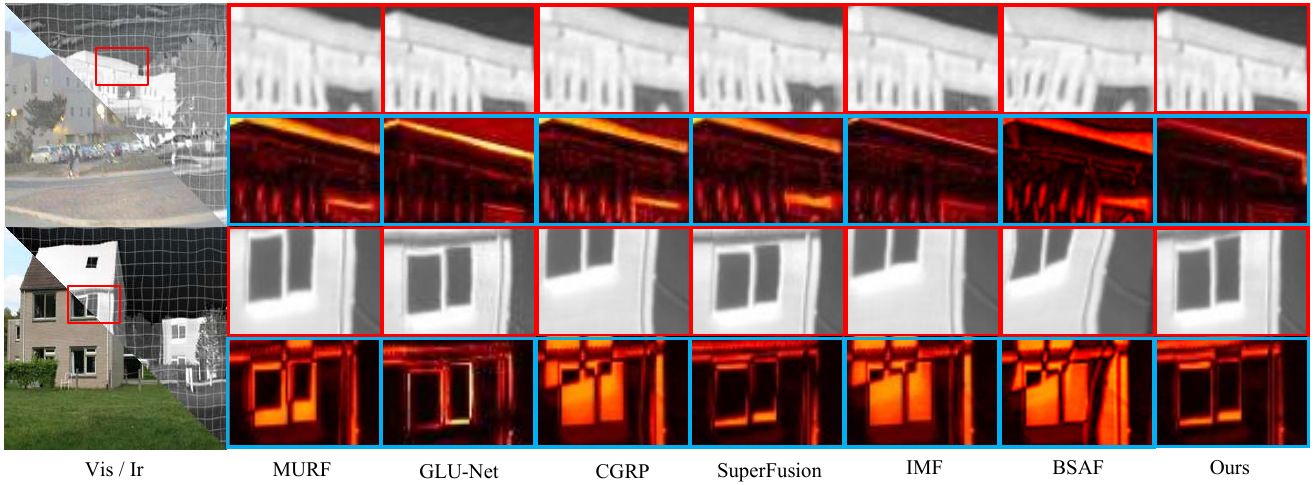}
    \caption{\textbf{Qualitative results of registration.} The blue boxes are the difference maps, where darker colors indicate smaller differences.}
    \label{compare_registration}
\end{figure*}

\section{Experiments}
\textbf{Implementation Details.}
We train our model on an NVIDIA A100 GPU with Adam optimizer for parameter updates. The initial learning rate is set to \(1e^{-3}\), employing an exponential decay strategy to refine the learning process over time. The training was executed with a patch size \(128 \times 128\) and a batch size of 16. 

\noindent \textbf{Datasets and Evaluation Metrics.}
We conduct infrared and visible image registration and fusion experiments on four datasets: RoadScene~\cite{xu2020u2fusion}, MSRS~\cite{tang2022piafusion}, M\(^{3}\)FD~\cite{liu2022target}, and TNO~\cite{toet2017tno}. Infrared images are randomly deformed to create misaligned image pairs. The RoadScene, MSRS, and M\(^{3}\)FD datasets are randomly split into 80\% for training and 20\% for testing, while the TNO dataset is used solely for testing due to its limited number of images.

For evaluation, we use four metrics for registration: normalized cross-correlation (NCC), root mean square error (RMSE), max square error (MAE), and median square error (MEE). For fusion, we use the correlation coefficient (CC), visual information fidelity (VIF), structural similarity index (SSIM), mean gradient (MG)~\cite{ma2019infrared}, and edge intensity (EI)~\cite{luo2017multi} for evaluation.

\begin{table*}[!t]
\centering
  \scriptsize
  \renewcommand\arraystretch{1.1} 
  \setlength{\tabcolsep}{0.8mm}
  \caption{\textbf{Quantitative comparison of IVIF.} The best results are in \textbf{bold}, while the second-best results are \underline{underlined}.}
  \resizebox{\textwidth}{!}{
  \begin{tabular}{lc|ccc|ccc|ccc|ccc}
    \toprule
    \multicolumn{2}{l|}{\cellcolor{gray!14}}     & \multicolumn{3}{c|}{\cellcolor{gray!14}RoadScene} & \multicolumn{3}{c|}{\cellcolor{gray!14}MSRS} & \multicolumn{3}{c|}{\cellcolor{gray!14}\(\text{M}^{3}\text{FD}\)} & \multicolumn{3}{c}{\cellcolor{gray!13}TNO}  \\
     \multicolumn{2}{c|}{\cellcolor{gray!14}Methods} & \cellcolor{gray!14}CC$\uparrow$ & \cellcolor{gray!14}SSIM $\uparrow$ & \cellcolor{gray!14}MG $\uparrow$ & \cellcolor{gray!14}CC $\uparrow$ & \cellcolor{gray!14}SSIM $\uparrow$ & \cellcolor{gray!14}MG $\uparrow$ & \cellcolor{gray!14}CC $\uparrow$ & \cellcolor{gray!14}SSIM $\uparrow$ & \cellcolor{gray!14}MG $\uparrow$ & \cellcolor{gray!14}CC $\uparrow$ & \cellcolor{gray!14}SSIM $\uparrow$ & \cellcolor{gray!14}MG $\uparrow$  \\
    \midrule
    DIDFuse~\cite{zhao2020didfuse}   & IJCAI'20   & 0.773   & 0.456  & 6.323  & 0.545  & 0.304  & 2.728  & 0.751  & 0.486  & \underline{3.739}  & 0.339  & 0.487  & 5.637   \\
    U2Fusion~\cite{xu2020u2fusion}   & TPAMI'20    & 0.752  & 0.464  & 4.975  & 0.565  & 0.577  & 2.278  & 0.777  & 0.674  & 2.891  & \underline{0.351} & 0.532 & 4.388    \\
    ReCoNet~\cite{huang2022reconet}   & ECCV'22    & 0.771  & 0.538  & 5.087  & 0.586  & 0.384  & \underline{4.546}  & 0.794  & 0.656  & 2.690  & 0.297 & 0.554 & 4.588   \\
    SuperFusion~\cite{tang2022superfusion}  & JAS'22      & \underline{0.785}  & \textbf{0.719}  & 5.551  & \underline{0.624}  & 0.608  & 3.528  & \underline{0.795}  & \textbf{0.792}  & 3.412  & 0.320 & \underline{0.602} & 4.111   \\
    MURF~\cite{xu2023murf}  & TPAMI'23     & 0.755  & 0.484  & 5.487  & 0.554  & 0.503  & 2.859  & 0.761  & 0.678  & 3.580  & 0.329 & 0.514 & 4.806    \\
    CDDFusion~\cite{zhao2023cddfuse}  & CVPR'23     & 0.760  & 0.456  & 6.000  & 0.585  & 0.592  & 3.626  & 0.784  & 0.655  & 3.223  & 0.320 & 0.507 & \textbf{6.100}  \\
    SegMIF~\cite{liu2023multi}  & ICCV'23     & 0.761  & 0.547  & 4.512  & 0.600  & 0.506  & 3.889  & 0.783  & 0.696  & 2.836  & 0.325 & 0.528 & 3.762  \\
    ReFusion~\cite{bai2024refusion}  & IJCV'24     & 0.735  & 0.468  & \underline{6.464}  & 0.584  & \underline{0.643}  & 3.413  & 0.763  & 0.739  & 3.528  & 0.292 & 0.546 & \underline{6.076}    \\
    MRFS~\cite{zhang2024mrfs}  & CVPR'24     & 0.762  & 0.576  & 5.885 & 0.591  & 0.625  & 2.347  & 0.748  & 0.707  & 2.623  & 0.234 & 0.407 & 4.297    \\
    SHIP~\cite{zheng2024probing}  & CVPR'24     & 0.717  & 0.480  & 6.042  & 0.580  & 0.617  & 3.636  & 0.726  & 0.724  & 3.618  & 0.301 & 0.551 & 5.547    \\
    BSAFusion~\cite{li2024bsafusion}  & AAAI'25     & 0.536  & 0.365  & 5.550  & 0.456  & 0.366  & 3.197  & 0.565  & 0.534  & 3.032  & 0.307 & 0.429 & 4.589    \\
    \midrule
    \textbf{Ours}  &\textbf{-}  & \textbf{0.792} & \underline{0.681} & \textbf{7.104} & \textbf{0.668} & \textbf{0.681} & \textbf{5.298} & \textbf{0.807} & \underline{0.760} & \textbf{3.948} & \textbf{0.474} & \textbf{0.647} & 5.248  \\
    \bottomrule
  \end{tabular}
  }
  \label{tab:fusion_quantitative_results}%
\end{table*}

\begin{figure*}[!t]
 \centering
 \includegraphics[width=1\textwidth]{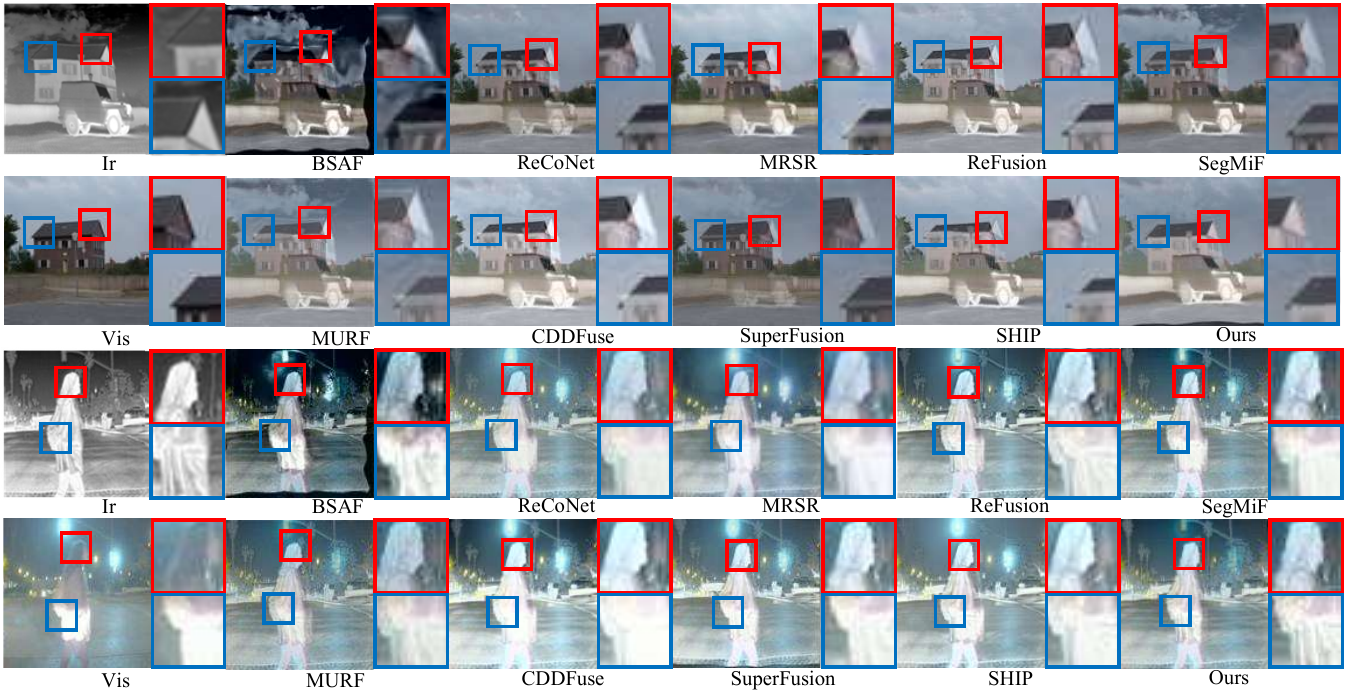}
    \caption{\textbf{Qualitative results of IVIF.}}
    \label{compare_fusion}
\end{figure*}

\subsection{Experiments on Infrared Image Registration}
We compare our method with six state-of-the-art approaches, including GLU-Net~\cite{truong2020glu}, CGRP~\cite{wang2022unsupervised}, SuperFusion~\cite{tang2022superfusion}, MURF~\cite{xu2023murf}, IMF~\cite{wang2024improving}, and BSAFusion~\cite{li2024bsafusion}. 
As shown in Table~\ref{tab:reg_quantitative_results}, our method achieves the best overall performance across all four test sets, and particularly dominates on M\(^{3}\)FD, yielding the highest \text{NCC} and consistently top-ranked \text{MEE}. These gains are visually confirmed in Figure~\ref{compare_registration}, where our results produce darker difference maps with fewer global and local misalignments. In contrast, CGRP often fails to recover the correct global transformation, while SuperFusion struggles with local non-rigid offsets. Benefiting from multi-scale iterative refinement of the deformation field, our approach handles both global and local deformations, delivering more accurate registration.

\subsection{Experiments on IVIF}
We compare our method with eleven IVIF approaches, including seven non-registration fusion methods (DIDFuse~\cite{zhao2020didfuse}, U2Fusion~\cite{xu2020u2fusion}, CDDFuse~\cite{zhao2023cddfuse}, SegMif~\cite{liu2023multi}, ReFusion~\cite{bai2024refusion}, MRFS~\cite{zhang2024mrfs}, SHIP~\cite{zheng2024probing}) and four registration-fusion methods (ReCoNet~\cite{huang2022reconet}, SuperFusion~\cite{tang2022superfusion}, MURF~\cite{xu2023murf}, and BSAFusion~\cite{li2024bsafusion}). Non-registration methods take visible and deformed infrared images as input, while registration-fusion methods operate on visible and registered infrared images. As reported in Table~\ref{tab:fusion_quantitative_results}, our method achieves the highest \text{CC} on all four datasets, ranking first on MSRS and consistently top-two on RoadScene and M\(^{3}\)FD. These quantitative gains are further reflected in the visual results in Figure~\ref{compare_fusion}. Non-registration methods suffer from noticeable artifacts caused by infrared deformation, whereas registration-fusion methods still exhibit residual distortions. Our approach produces visually coherent fusion results with enhanced thermal saliency and spatial consistency, benefiting from the proposed Dual-branch Spatial-Frequency Fusion (DSFF) module.

\begin{table}[!pt]
  \centering
  \small 
  \setlength{\tabcolsep}{1.9mm} 
  \renewcommand{\arraystretch}{1.2} 
  \caption{Quantitative ablation study of image registration.}
  \begin{tabular}{ccc|ccc}
    \toprule
    \rowcolor{gray!6}\multicolumn{1}{c}{FRB} & \multicolumn{1}{c}{MSF} & \multicolumn{1}{c}{$\mathcal{L}_{uncertainty}$} & \multicolumn{1}{c}{NCC}$\uparrow$ & \multicolumn{1}{c}{RMSE}$\downarrow$ & \multicolumn{1}{c}{MEE}$\downarrow$  \\
    \midrule
       & \checkmark & \checkmark    &0.752   &5.791  &5.682  \\
     \checkmark &    &  \checkmark &0.831  & 5.458  &5.370 \\
     \checkmark & \checkmark  &  &0.770  &5.716 &5.640 \\
     \checkmark     & \checkmark    &\checkmark  & 0.948  & 5.073  & 5.043\\
    \bottomrule
  \end{tabular}
  \label{tableablation_reg}
\end{table}

\begin{figure}[!t]
 \centering
 \includegraphics[width=\columnwidth]{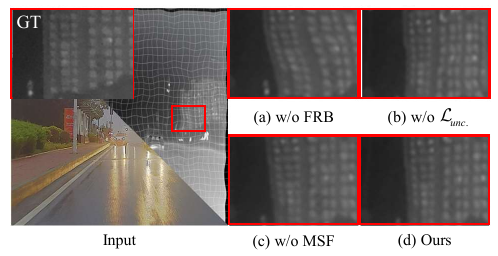}
    \caption{Qualitative ablation study of image registration.}
    \label{ablation_registration}
\end{figure}

\subsection{Ablation Studies}
Tables~\ref{tableablation_reg} and~\ref{tableablation_fusion} show that removing any key component consistently degrades performance. For registration, discarding the GRU-based FRB notably reduces NCC/RMSE and introduces visible misalignment (Figure~\ref{ablation_registration}), validating the benefit of iterative refinement. Removing uncertainty modeling (either $\mathcal{L}_{uncertainty}$ or MSF) further harms stability, especially in blurry regions. For fusion, omitting $\mathcal{L}^{r}_{freq}$ weakens frequency-domain consistency and thermal saliency, while removing FDF or SDF degrades fusion quality by losing thermal cues or structural details, respectively (Tables~\ref{tableablation_fusion}).

\begin{table}[!pt]
  \centering
  \small 
  \setlength{\tabcolsep}{3.3mm} 
  \renewcommand{\arraystretch}{1.2} 
  \caption{Quantitative ablation study of IVIF.}
  \begin{tabular}{ccc|ccc}
    \toprule
    \rowcolor{gray!6}\multicolumn{1}{c}{$\mathcal{L}^{r}_{freq}$} & \multicolumn{1}{c}{FDF} & \multicolumn{1}{c}{SDF} & \multicolumn{1}{c}{CC}$\uparrow$ & \multicolumn{1}{c}{SSIM}$\uparrow$ & \multicolumn{1}{c}{MG}$\uparrow$  \\
    \midrule
     &\checkmark  & \checkmark     &0.730  &0.604  &6.550  \\
      \checkmark &    & \checkmark  &0.524  &0.577  &5.483 \\
     \checkmark     & \checkmark  &  &0.663  &0.505  &5.336  \\
     \checkmark &\checkmark  & \checkmark     & 0.792 & 0.681 &7.104 \\
    \bottomrule
  \end{tabular}
  \label{tableablation_fusion}
\end{table}



\section{Conclusion}
We propose the Spatial-Frequency Registration and Fusion (SFRF) framework to address the misalignment and error accumulation in multi-modal image fusion. By incorporating uncertainty estimation and infrared thermal radiation distribution consistency, SFRF improves the robustness of registration and fusion across both spatial and frequency domains. Our method provides a promising solution for accurate and robust infrared and visible image fusion.

\textbf{Broader Impacts.} 
Our work enhances infrared-visible image fusion by reducing error accumulation and preserving thermal consistency, which may improve registration robustness in critical applications such as surveillance and remote sensing.